\documentclass[10pt,twocolumn,letterpaper]{article}

\usepackage{iccv}
\usepackage{times}
\usepackage{epsfig}
\usepackage{graphicx,multirow,booktabs}
\usepackage{amsmath}
\usepackage{amssymb}

\usepackage[breaklinks=true,bookmarks=false]{hyperref}

\iccvfinalcopy 

\def\iccvPaperID{****} 

\ificcvfinal\fi

\begin{document}

\title{DAN: A Deformation-Aware Network for Consecutive Biomedical Image Interpolation}

\author{Zejin Wang\qquad Guoqing Li \qquad Xi Chen\qquad Hua Han\\
Institute of Automation, Chinese Academy of Sciences \\
School of Artificial Intelligence, Chinese Academy of Sciences University \\
National Laboratory of Pattern Recognition, CASIA \\
Center of Excellence in Brain Science and Intelligence Technology, CAS
{\tt\small wangzejin2018@ia.ac.cn}
}

\maketitle
\ificcvfinal\thispagestyle{empty}\fi

\begin{abstract}
   The continuity of biological tissue between consecutive biomedical images makes it possible for the video interpolation algorithm, to recover large area defects and tears that are common in biomedical images. However, noise and blur differences, large deformation, and drift between biomedical images, make the task challenging. To address the problem, this paper introduces a deformation-aware network to synthesize each pixel in accordance with the continuity of biological tissue. First, we develop a deformation-aware layer for consecutive biomedical images interpolation that implicitly adopting global perceptual deformation. Second, we present an adaptive style-balance loss to take the style differences of consecutive biomedical images such as blur and noise into consideration. Guided by the deformation-aware module, we synthesize each pixel from a global domain adaptively which further improves the performance of pixel synthesis. Quantitative and qualitative experiments on the benchmark dataset show that the proposed method is superior to the state-of-the-art approaches.
\end{abstract}

\section{Introduction}

Consecutive biomedical image interpolation is important in biomedical image analysis. Mechanical damage in the sample preparation process and the destruction of biological tissues during electron microscope imaging can cause large-area defects in consecutive biomedical images. At present, the preliminary work in the field of sequence slice insertion is little~\cite{nguyen2019weakly}, and the image restoration methods are not effective when dealing with large area defects. However, interpolation methods can predict non-defective intermediate frames, which can replace intermediate frames with large area defects. Besides, intermediate images contribute to improve registration results with sudden and large structural changes, improve semantic segmentation accuracy, and ensure 3D reconstruction continuity. \par
For the natural image, frame interpolation methods process occlusion, large motion, adopting depth maps, optical flow, and local interpolation kernels. Depth~\cite{eigen2014depth,eigen2015predicting,wang2015towards,liu2015learning,roy2016monocular,chen2016single,kuznietsov2017semi,fu2018deep} is one of the key visual information to understand the 3D geometry of a scene and has been widely exploited in several nature scenes. However, each pixel of the electron microscope image is at the same depth, so depth estimation is not required. The ultimate goal is to synthesize high-quality intermediate missing frames and optical flow estimation~\cite{dosovitskiy2015flownet,ilg2017flownet,ranjan2017optical} is used only as an intermediate step. Kernel estimation based methods~\cite{niklaus2017adaptive,niklaus2017separable} combine motion estimation and frame synthesis into a single convolution step. Large estimation kernels are necessary if large deformation occurs. The memory consumption increases quadratically with the kernel size and thus limits the large motion to be handled. Besides, traditional convolution used in kernel estimation is unable to build a global relationship. As for feature extractor, kernel-based methods adopt the U-Net structure to extract the hierarchical features of the input frames. The pooling process of U-Net can damage context information, making it difficult for intermediate frame synthesis. \par
In this work, we propose to implicitly detect the deformation by utilizing the deformation-aware layer for consecutive biomedical images interpolation. The proposed deformation-aware layer exploits a two-level sparse self-attention mechanism to build global dependency for the output frame. This means that each pixel of the intermediate frame takes into account the global deformation between the input frames. Specifically, we first learn hierarchical features from the two input frames via the residual dense network~\cite{zhang2018residual}. To better preserve timing information between two consecutive frames, the siamese structure is adopted for feature extraction. Then, the proposed deformation-aware layer processes the input feature maps and synthesizes the warped frames. In contrast to the kernel estimation layer, the proposed deformation-aware layer generates intermediate frames with clearer boundaries due to the global perceptual deformation. Extensive experiments on multiple types of datasets, produced by scanning electron microscopy, indicate that the proposed method performs against existing consecutive biomedical image interpolation methods.

\section{Related Work}

\begin{figure*}[t]
\begin{center}
\includegraphics[width=\textwidth]{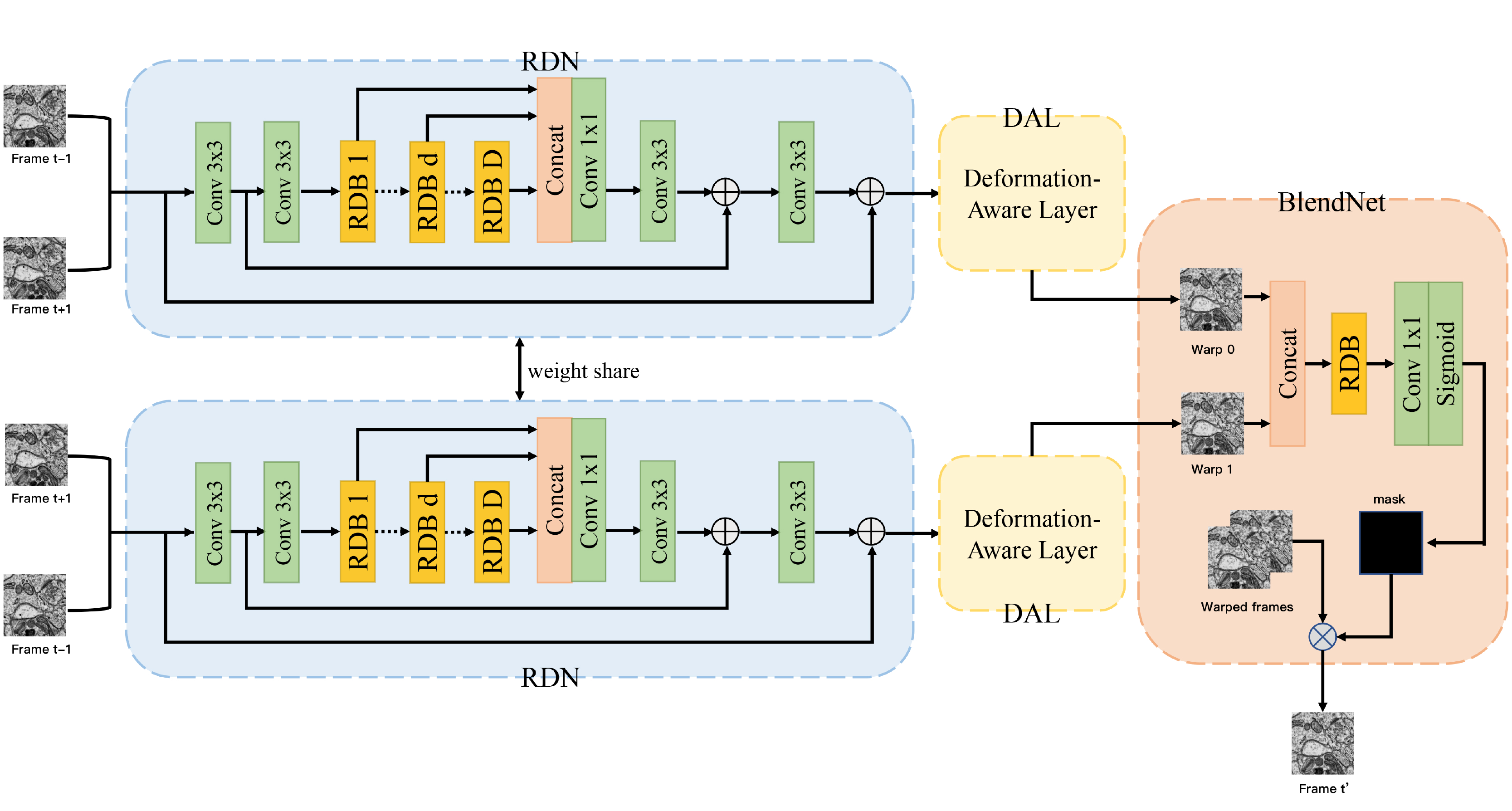}
\end{center}
   \caption{Overview of the DAN algorithm, which includes the siamese residual dense network, deformation-aware layers, and blendnet.} 
\label{fig:overview}
\end{figure*}

\section{Method}
An overview of the proposed deformation-aware interpolation algorithm is shown in Fig ~\ref{fig:overview}, which is primarily based on siamese residual dense network, deformation-aware layer and blendnet. Given two input frames $\mathbf{I}_{t-1}$ and $\mathbf{I}_{t+1}$, the goal is to synthesize an intermediate frame $\hat{\mathbf{I}}_{t}$. We first encode the feature maps, denoted by $\mathbf{F}_{t-1\to t+1}$ and $\mathbf{F}_{t+1\to t-1}$ , through siamese residual dense network. Then, the proposed deformation-aware layer synthesizes the warped frames $\mathbf{warp}_{0}$ and $\mathbf{warp}_{1}$ based on $\mathbf{F}_{t-1\to t+1}$ and $\mathbf{F}_{t+1\to t-1}$. After obtaining the warped frames, the blendnet generates the interpolated frame $\hat{\mathbf{I}}_{t}$ by linearly fusing. \par

\subsection{Siamese Residual Dense Network}
\label{sec:siam}
To preserve the structured information when generating the corresponding hierarchical features of the input frames, we utilize the residual dense network as the basic feature extractor. As shown in Fig ~\ref{fig:overview}, residual dense network(RDN) mainly consists of three parts: shallow feature extraction net(SFENet), residual dense blocks (RDBs), and finally dense feature fusion (DFF). Besides, the frame interpolation task requires two consecutive frames as input to generate intermediate frames. Here, the siamese structure is adopted as illustrated in Fig~\ref{fig:overview}, which preserves the temporal information between consecutive frames during generating hierarchical features and contributes to decrease the computational consumption. \par

\begin{figure*}[t]
\begin{center}
\includegraphics[width=\textwidth]{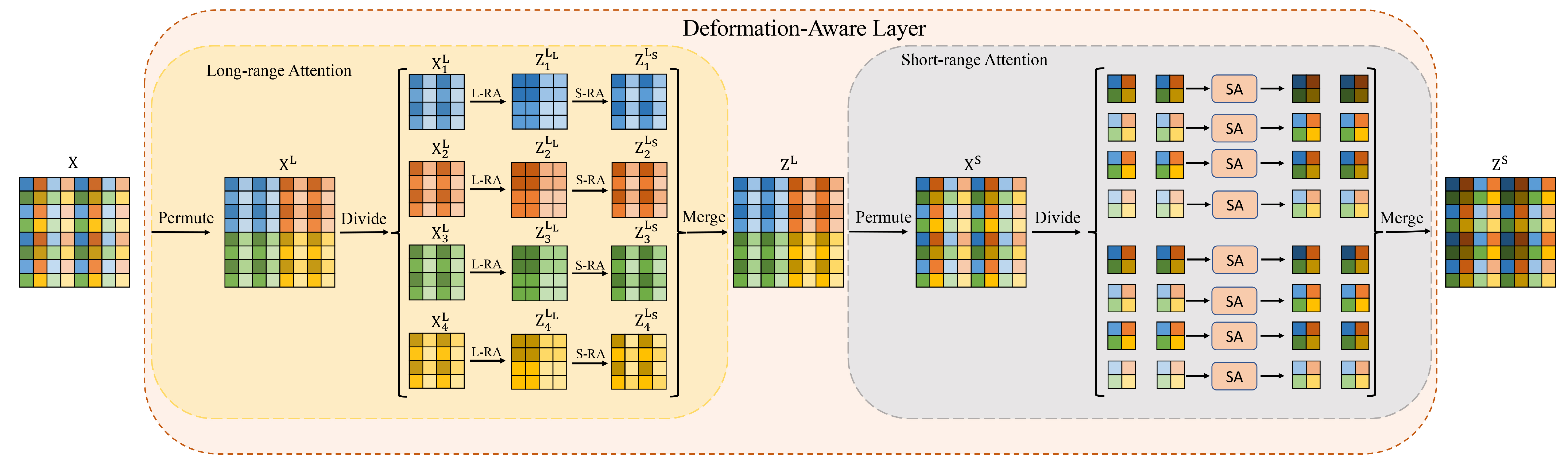}
\end{center}
   \caption{Illustration of Deformation-Aware layer. L-RA and S-RA denote long-range attention and short-range attention, respectively.}
\label{fig:attn}
\end{figure*}

\subsection{Deformation-Aware Layer}
\label{sec:da}
After extracting hierarchical features through the siamese residual dense network, the proposed Deformation-Aware layer based on multi-level sparse self-attention replaces the kernel estimation. The key of the proposed multi-level sparse self-attention lies in the multi-level decomposition of the original dense affinity matrix $\mathbf{A}$, each time decomposing the dense affinity matrix $\mathbf{A}$ into the product of two sparse block affinity matrices $\mathbf{A}^{L}$ and $\mathbf{A}^{S}$. By combining multi-level decomposition, long-range attention, and short-range attention, pixels of each position can be synthesized from the information of all input positions. We demonstrate how to estimate the long-range attention matrix $\mathbf{A}^{L}$ or the short-range attention matrix $\mathbf{A}^{S}$ and perform multi-level decomposition in Figure~\ref{fig:attn}.\par
\noindent\textbf{Long-range Attention.} Long-range attention apply the self-attention on the subsets of positions that satisfy long spatial interval distances.
As shown in Fig~\ref{fig:attn}, a permutation is first adopted on the input feature map $\mathbf{X}$ to compute $\mathbf{X}^{L}=Permute(\mathbf{X})$. Then, $\mathbf{X}^{L}$ is divided into $\mathcal{P}$ parts and each part contains $\mathcal{Q}$ adjacent positions($N=\mathcal{P}\times \mathcal{Q}$). Here, 
\begin{equation}
    \mathbf{X}^{L}=[{\mathbf{X}_{1}^{L}},{\mathbf{X}_{2}^{L}},...,{\mathbf{X}_{\mathcal{P}}^{L}}],
\end{equation}

\begin{equation}
    \mathbf{A}_{p}^{L}=softmax(\frac{(\mathbf{W_{\mathit{f}}} \mathbf{X}_{p}^{L})^\top(\mathbf{W_{\mathit{g}}} \mathbf{X}_{p}^{L})}{\sqrt{d}}),
\end{equation}

\begin{equation}
    \mathbf{Z}_{p}^{L}=\mathbf{W_{\mathit{v}}}[(\mathbf{W_{\mathit{h}}} \mathbf{X}_{p}^{L})\mathbf{A}_{p}^{L}],
\end{equation}

\begin{equation}
    \label{ZL}
    \mathbf{Z}^{L}=[{\mathbf{Z}_{1}^{L}},{\mathbf{Z}_{2}^{L}},...,{\mathbf{Z}_{\mathcal{P}}^{L}}],
\end{equation}
\noindent where each $\mathbf{X}_{p}^{L} \in \mathbb{R}^{\mathcal{C} \times \mathcal{Q}}$ is a subset of $\mathbf{X}^{L}$, $\mathbf{A}_{p}^{L} \in \mathbb{R}^{\mathcal{Q} \times \mathcal{Q}}$ is the sparse affinity matrix based on all the positions from input feature map $\mathbf{X}_{p}^{L}$ and $\mathbf{Z}_{p}^{L} \in \mathbb{R}^{\mathcal{C} \times \mathcal{Q}}$ is the updated output feature map based on input feature map $\mathbf{X}_{p}^{L}$. All other parameters including $\mathbf{W}_{\mathit{f}},\mathbf{W}_{\mathit{g}},\mathbf{W}_{\mathit{v}},\mathbf{W}_{\mathit{h}},d$ are the same as self-attention section. Finally, all the $\mathbf{Z}_{p}^{L}$ is merged to acquire the output feature map $\mathbf{Z}^{L}$ in equation ~\ref{ZL}. From the equations above, we demonstrate the actual affinity matrix of long-range attention as below,
 \begin{equation}
    \mathbf{A}^{L}=
            diag(\mathbf{A}_{1}^{L}, \mathbf{A}_{2}^{L},...,\mathbf{A}_{\mathcal{P}}^{L}),
\end{equation}
\noindent The equation shows that only the small affinity blocks in the diagonal are non-zero.\par
\noindent\textbf{Short-range Attention.}      
Short-range attention apply the self-attention on the subsets of positions that satisfy short spatial interval distances. The decomposition principle is similar to the long-range attention mechanism. 

\noindent\textbf{Multi-level Decomposition.}
The combination of long-range attention and short-range attention can effectively model global dependence. However, the computation of the small affinity matrix $\mathbf{A}_{p}^{L}$ from long-range attention is still not very efficient. We continue to decompose the sub-feature map $\mathbf{X}_{p}^{L}$. Here, we only perform two-level decomposition. As illustrated in Fig~\ref{fig:attn}, we first adopt a permutation on the input feature map $\mathbf{X}$ to compute $\mathbf{X}^{L}$ and divide $\mathbf{X}^{L}$ into $\mathcal{P}$ parts. Second, we apply a permutation on the input sub-feature map $\mathbf{X}_{p}^{L}$ to compute $\mathbf{X'}_{p}^{L}=Permute(\mathbf{X}_{p}^{L})=[{\mathbf{X'}_{p1}^{L}},{\mathbf{X'}_{p2}^{L}},...,{\mathbf{X'}_{p\mathcal{P'}}^{L}}],N'=\mathcal{P' \times Q'}$. Parameters here are similar to long-range attention. Then, we repeat the previous long-range attention and short-range attention steps in sequence, to calculate $\mathbf{Z}_{p}^{L_{L}}$ and $\mathbf{Z}_{p}^{L_{S}}$. After acquiring the updated output feature map based on the input sub-feature map $\mathbf{X}_{p}^{L}$, we merge all the $\mathbf{Z}_{p}^{L_{S}}$ to acquire the output feature map $\mathbf{Z}^{L}$. Finally, the output feature map $\mathbf{Z}^{S}$ can be obtained through performing short-range attention on $\mathbf{Z}^{L}$ directly. \par

\noindent\textbf{Complexity.}
For the implementation of the self-attention, we directly utilize the open-source code~\cite{zhang2018self}. Given the input feature map of size $H \times W \times C$, the complexity of interlaced sparse self-attention~\cite{huang2019interlaced} can be minimized to $\mathcal{O}(4HWC^{2}/k+3(HW)^{\frac{3}{2}}C/k)$, and the complexity of the proposed method can be minimized to $\mathcal{O}(12HWC^{2}/k+6(HW)^{\frac{4}{3}}C/k)$. Thus, our method has a significantly lower computational complexity than the first one. \par

\noindent{\bf{Loss Function.}}
Biomedical images are quite different from general natural images, since biomedical images are imaged through the electron microscope. These images have the characteristics of rich in noise and different degrees of blur, which determine that general loss functions are not suitable for consecutive biomedical image interpolation task. Thus, the loss function for training the proposed network is a combination of \textit{style reconstruction loss}, \textit{feature reconstruction loss}, and \textit{pixel-wise loss}. Specifically, the total loss in the proposed algorithm is $
    \mathcal{L}_{total}= \alpha\cdot\mathcal{L}_{s} + \beta\cdot\mathcal{L}_{f} + \gamma\cdot\mathcal{L}_{1}
$,
where the scalar $\alpha, \beta, \gamma$ are the trade off weight, and the constant $\alpha$ is $1e6$, the constant $\beta = \gamma = 1$,  $\mathcal{L}_{s}, \mathcal{L}_{f}, \mathcal{L}_{1}$ denote \textit{style reconstruction loss}, \textit{feature reconstruction loss} and \textit{pixel-wise loss}, respectively. \par

\subsection{Implementation Details}
Below we discuss implementation details with respect to training dataset, training strategy, and data augmentation. \par
\noindent{\bf{Training Dataset.}}
We evaluate the proposed approach on two major types of biomedical images: ssTEM images and ATUM images. Here, we use the CREMI\footnote{https://cremi.org/} dataset, provide by MICCAI 2016 Challenge as ssTEM images. To avoid limitations of data types, we also use our mouse cell datasets based on ATUM with z-axis resolutions of 8 nanometers as ATUM images. Each dataset adopts a triplet as a sample for training, where each triplet contains 3 consecutive biomedical images with a resolution of $512\times512$ pixels. To eliminate the brightness inconsistency of the electron microscope image, histogram specification is performed on each dataset, which contributes to improving the robustness of the algorithm. \par

\noindent{\bf{Training Strategy.}}
The proposed models are optimized using the Adam~\cite{kingma2014adam} with the $\beta_{1}$ of $0.9$ and $\beta_{2}$ of $0.999$. We set the batch size to 3 with synchronized batch normalization. The initial learning rates of the proposed network are set to 1e-3. We train the entire model for 30 epochs and then reduce the learning rate by a factor of 0.1 and fine-tune the entire model for another 20 epochs. Training requires approximately 3 days to converge on one Tesla K80 GPU. The whole DAN framework is implemented using PyTorch. \par

\noindent{\bf{Data augmentation.}}
For data augmentation, We randomly flipped the cropped patches horizontally or vertically and randomly swap their temporal order, for all datasets. All input images are randomly cropped to $512\times 512$.\par

\section{Experiments}
In this section, we first conduct ablation study to analyze the contribution of the proposed feature extractor,deformation-aware layer and loss function. Then, we compare the proposed model with state-of-the-art algorithms on different types of biomedical datasets. The average Interpolation Error(IE), PSNR and SSIM are computed for comparisons. Lower IEs indicate better performance.\par

\begin{figure*}[t]
\includegraphics[width=\textwidth]{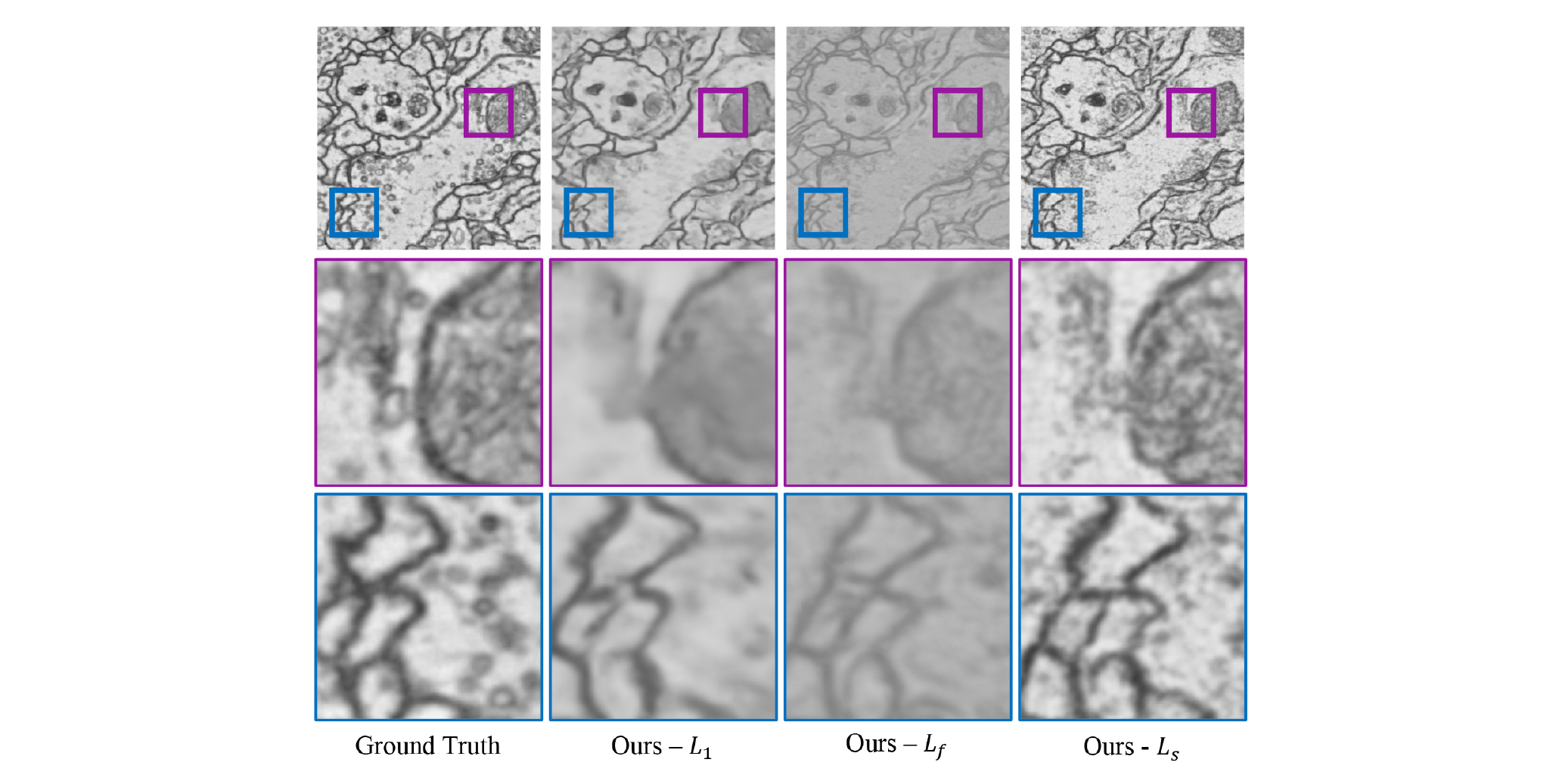}
\caption{The effect of loss functions.} 
\label{fig:loss}
\end{figure*}

\noindent{\bf{Loss functions.}}
The proposed method incorporates three types of loss functions: pixel-wise loss $\mathcal{L}_{1}$, feature reconstruction loss $\mathcal{L}_{f}$ and style reconstruction loss $\mathcal{L}_{s}$. To reflect their respective effects, three different loss functions are adopted to train the proposed network. The first one only applies $\mathcal{L}_{1}$ loss and represents this network as "$L_{1}$". The second one applies both $\mathcal{L}_{1}$ loss and $\mathcal{L}_{f}$ loss in a linear combination and represents this network as "$L_{f}$". The third one applies $\mathcal{L}_{1}$ loss, $\mathcal{L}_{f}$ loss and $\mathcal{L}_{s}$ loss in a linear combination and represents this network as "$L_{s}$". As shown in Fig \ref{fig:loss}, the last "$L_{s}$" leads to the best visual quality and rich texture information. Results generated by "$L_{s}$" are visually pleasing with more high-frequency details. In addition, the style of the images is almost the same as the ground truth. As a result, the proposed network adopts this scheme as the loss function.

\subsection{Model Analysis}
We analyze the contribution of the two key components in the proposed model: the siamese residual dense network and the deformation-aware layer. \par

\noindent{\bf{Siamese residual dense network.}}
To analyze the effectiveness of the siamese residual dense network(SRDN), this feature extractor is compared with other famous feature extractors, including the U-Net feature extractor(U-Net), the siamese U-Net feature extractor(SU-Net), and the residual dense network(RDN) on the CREMI datasets and ps8nm mouse dataset. As shown in Table \ref{extractor}, the proposed SRDN feature extractor outperforms previous state-of-the-art feature extractors, obtaining almost the best performance on PSNR, SSIM and IE.

\noindent{\bf{Deformation-aware layer.}}
To analyze the effectiveness of the proposed deformation-aware layer, the feature extractor adopts a siamese residual dense network. After the feature extractor, we append the classic kernel estimation layer, the state-of-the-art interlaced sparse self-attention layer and the proposed deformation-aware layer, respectively. As shown in Table \ref{synthesis}, the proposed deformation-aware layer(DAL) shows a substantial improvement  on the CREMI and ps8nm mouse datasets, against both kernel estimation layer(KEL) and interlaced sparse self-attention layer(SSA).

\begin{table*}[t]
\centering 
    \begin{tabular}[b]{*{11}{c}}
    \toprule
    \multirow{2}*{Extractor} & 
    \multicolumn{2}{c}{CREMI A} & 
    \multicolumn{2}{c}{CREMI B} &
    \multicolumn{3}{c}{CREMI C} &
    \multicolumn{3}{c}{ps8nm mouse} \\ 
    \cmidrule(lr){2-3} \cmidrule(lr){4-5}  \cmidrule(lr){6-8}  \cmidrule(lr){9-11} & 
    \multicolumn{1}{c}{PSNR} &
    \multicolumn{1}{c}{SSIM} &
    \multicolumn{1}{c}{PSNR} &
    \multicolumn{1}{c}{SSIM} &
    \multicolumn{1}{c}{PSNR} &
    \multicolumn{1}{c}{SSIM} &
    \multicolumn{1}{c}{IE} &
    \multicolumn{1}{c}{PSNR} &
    \multicolumn{1}{c}{SSIM} &
    \multicolumn{1}{c}{IE}  \\
    \midrule
    U-Net~\cite{ronneberger2015u} & 18.10 & 0.4295  & 16.75 & 0.3549 &  16.26 & 0.3542 & 28.82 & 14.59 & 0.2219 & 36.55 \\
    SU-Net & 18.12 & 0.4352  & 16.71 & 0.3417 & 16.18 & 0.3280 & 29.71 & 14.75 & 0.2089 & 35.99\\
    RDN~\cite{zhang2018residual} & 17.73 & \textbf{0.4448}  & 16.59 & 0.3521  & 16.44 & 0.3451 & 28.75 & 14.85 & 0.2195 & 35.53 \\
    SRDN(ours) & \textbf{18.26} & 0.4374  & \textbf{16.79} & \textbf{0.3712}  & \textbf{16.46} & \textbf{0.3575} & \textbf{28.38} & \textbf{15.04} & \textbf{0.2156} & \textbf{34.81}\\
    \bottomrule
    \end{tabular}
    \caption{Results on hierarchical features.}
    \label{extractor}
\end{table*}

\begin{table*}[t]
\centering 
    \begin{tabular}[b]{*{11}{c}}
    \toprule
    \multirow{2}*{Synthesis} & 
    \multicolumn{2}{c}{CREMI A} & 
    \multicolumn{2}{c}{CREMI B} &
    \multicolumn{3}{c}{CREMI C} &
    \multicolumn{3}{c}{ps8nm mouse} \\ 
    \cmidrule(lr){2-3} \cmidrule(lr){4-5}  \cmidrule(lr){6-8}  \cmidrule(lr){9-11} & 
    \multicolumn{1}{c}{PSNR} &
    \multicolumn{1}{c}{SSIM} &
    \multicolumn{1}{c}{PSNR} &
    \multicolumn{1}{c}{SSIM} &
    \multicolumn{1}{c}{PSNR} &
    \multicolumn{1}{c}{SSIM} &
    \multicolumn{1}{c}{IE} &
    \multicolumn{1}{c}{PSNR} &
    \multicolumn{1}{c}{SSIM} &
    \multicolumn{1}{c}{IE}  \\
    \midrule
    KEL~\cite{niklaus2017separable} & 17.59 & 0.4354  & 15.88 & 0.3415 &  16.08 & 0.3506 & 30.82 & 13.46 & 0.1867 & 42.62\\
    SSA~\cite{huang2019interlaced} & 18.12 & 0.4292  & 16.41 & 0.3425 &  16.33 & 0.3357 & 28.91 & 14.86 & \textbf{0.2261} & 35.41\\
    DAL(Ours) & \textbf{18.26} & \textbf{0.4374}  & \textbf{16.79} & \textbf{0.3712}  & \textbf{16.46} & \textbf{0.3575} & \textbf{28.38} & \textbf{15.04} & 0.2156 & \textbf{34.81}\\
    \bottomrule
    \end{tabular}
    \caption{Effects on Deformation-aware layer.} 
    \label{synthesis}
\end{table*}

\begin{table*}[t]
\centering 
    \begin{tabular}[b]{*{13}{c}}
    \toprule
    \multirow{2}*{Methods} & 
    \multicolumn{2}{c}{CREMI A} & 
    \multicolumn{2}{c}{CREMI B} &
    \multicolumn{3}{c}{CREMI C} &
    \multicolumn{3}{c}{ps8nm mouse} \\ 
    \cmidrule(lr){2-3} \cmidrule(lr){4-5}  \cmidrule(lr){6-8}  \cmidrule(lr){9-11} & 
    \multicolumn{1}{c}{PSNR} &
    \multicolumn{1}{c}{SSIM} &
    \multicolumn{1}{c}{PSNR} &
    \multicolumn{1}{c}{SSIM} &
    \multicolumn{1}{c}{PSNR} &
    \multicolumn{1}{c}{SSIM} &
    \multicolumn{1}{c}{IE} &
    \multicolumn{1}{c}{PSNR} &
    \multicolumn{1}{c}{SSIM} &
    \multicolumn{1}{c}{IE}  \\
    \midrule
    SepConv-L$_{s}$~\cite{niklaus2017separable} & 17.52 & 0.4095  & 16.32 & 0.3522  & 16.07 & 0.3454 & 29.82 & 12.56 & 0.1573 & 48.03  \\
     DAIN-L$_{s}$~\cite{DAIN} & 16.78 & 0.4264  & 15.67 & 0.3460  & 15.24 & 0.3210 & 33.59 & 13.06 & 0.1973 & 44.26
     \\
    DAN(Ours) & \textbf{18.26} & \textbf{0.4374}  & \textbf{16.79} & \textbf{0.3712}  & \textbf{16.46} & \textbf{0.3575} & \textbf{28.38} & \textbf{15.04} & \textbf{0.2156} & \textbf{34.81} \\
    \bottomrule
    \end{tabular}
    \caption{Quantitative comparisons on CREMI and ps8nm mouse.} 
    \label{comparisons}
\end{table*}

\begin{figure*}[h]
\includegraphics[width=\textwidth]{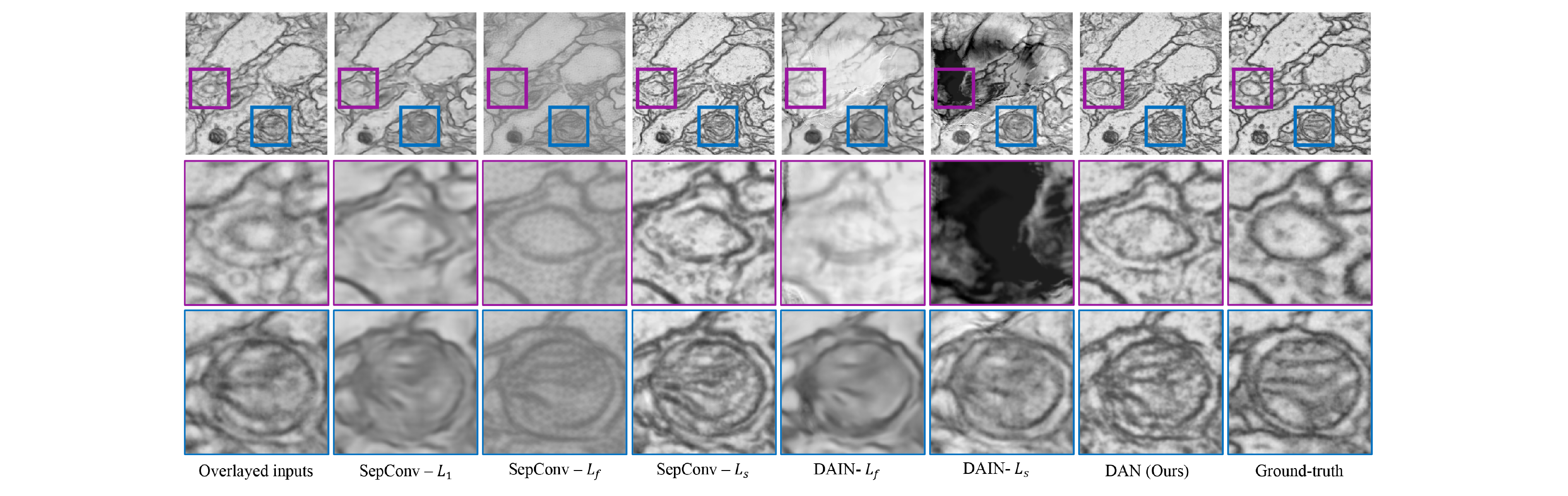}
\caption{Visual comparisons on the CREMI B.} 
\label{fig:visual}
\end{figure*}

\subsection{Comparisons with State-of-the-arts}
We evaluate the proposed DAN against the following CNN-based frame interpolation methods: SepConv-L$_{1}$~\cite{niklaus2017separable}, SepConv-L$_{f}$~\cite{niklaus2017separable},
SepConv-L$_{s}$~\cite{niklaus2017separable}, DAIN-L$_{1}$~\cite{DAIN} and DAIN-L$_{s}$~\cite{DAIN}, in terms of PSNR, SSIM and IE. In Table \ref{comparisons}, we provide quantitative performances on the CREMI A, CREMI B, CREMI C, and ps8nm mouse dataset. Our approach performs favorably against all the compared methods for all the datasets, especially on the ps8nm mouse dataset with a 2.48dB gain over SepConv-L$_{s}$~\cite{niklaus2017separable} in terms of PSNR. In Fig~\ref{fig:visual}, the DAIN-L$_{1}$~\cite{DAIN} and DAIN-L$_{s}$~\cite{DAIN} cannot handle the large deformation well and thus produce ghosting and broken results. The SepConv-L$_{1}$~\cite{niklaus2017separable} and SepConv-L$_{f}$~\cite{niklaus2017separable} methods generate blurred results on both membrane structure and mitochondria. The result generated by SepConv-L$_{s}$~\cite{niklaus2017separable} is more similar to the first frame than the intermediate frame, especially around mitochondria. In contrast, the proposed method handles large deformation well and generates clearer results.

\section{Conclusion}
In this paper, we propose a novel deformation-aware consecutive biomedical image interpolation algorithm, which combines motion estimation and frame synthesis into a single process using a deformation-aware layer. The proposed deformation-aware layer implicitly detects the large deformations using the self-attention information and synthesize each pixel by effectively establishing long-range dependencies from input frames. Furthermore, we exploit the siamese residual dense network as the feature extractor to learn hierarchical features and reduce the parameters. The experiments show that the proposed approach compares favorably to state-of-the-art interpolation methods both quantitatively and qualitatively and generates high-quality frame synthesis results.\par
\noindent{\bf{Acknowledgment.}} 
This work was supported in part by *** and ***.

{\small
\bibliographystyle{unsrt}

\bibliography{egbib}

\begin{thebibliography}{10}
\providecommand{\url}[1]{\texttt{#1}}
\providecommand{\urlprefix}{URL }
\providecommand{\doi}[1]{https://doi.org/#1}

\bibitem{DAIN}
Bao, W., Lai, W.S., Ma, C., Zhang, X., Gao, Z., Yang, M.H.: Depth-aware video
  frame interpolation. In: CVPR (2019)

\bibitem{chen2016single}
Chen, W., Fu, Z., Yang, D., Deng, J.: Single-image depth perception in the
  wild. In: NIPS (2016)

\bibitem{dosovitskiy2015flownet}
Dosovitskiy, A., Fischer, P., Ilg, E., Hausser, P., Hazirbas, C., Golkov, V.,
  Van Der~Smagt, P., Cremers, D., Brox, T.: Flownet: Learning optical flow with
  convolutional networks. In: ICCV (2015)

\bibitem{eigen2015predicting}
Eigen, D., Fergus, R.: Predicting depth, surface normals and semantic labels
  with a common multi-scale convolutional architecture. In: CVPR (2015)

\bibitem{eigen2014depth}
Eigen, D., Puhrsch, C., Fergus, R.: Depth map prediction from a single image
  using a multi-scale deep network. In: NIPS (2014)

\bibitem{fu2018deep}
Fu, H., Gong, M., Wang, C., Batmanghelich, K., Tao, D.: Deep ordinal regression
  network for monocular depth estimation. In: CVPR (2018)

\bibitem{huang2019interlaced}
Huang, L., Yuan, Y., Guo, J., Zhang, C., Chen, X., Wang, J.: Interlaced sparse
  self-attention for semantic segmentation. arXiv:1907.12273  (2019)

\bibitem{ilg2017flownet}
Ilg, E., Mayer, N., Saikia, T., Keuper, M., Dosovitskiy, A., Brox, T.: Flownet
  2.0: Evolution of optical flow estimation with deep networks. In: CVPR (2017)

\bibitem{kingma2014adam}
Kingma, D.P., Ba, J.: Adam: A method for stochastic optimization.
  arXiv:1412.6980  (2014)

\bibitem{kuznietsov2017semi}
Kuznietsov, Y., Stuckler, J., Leibe, B.: Semi-supervised deep learning for
  monocular depth map prediction. In: CVPR (2017)

\bibitem{liu2015learning}
Liu, F., Shen, C., Lin, G., Reid, I.: Learning depth from single monocular
  images using deep convolutional neural fields. NIPS  (2015)

\bibitem{nguyen2019weakly}
Nguyen-Duc, T., Yoo, I., Thomas, L., Kuan, A., Lee, W.c., Jeong, W.K.: Weakly
  supervised learning in deformable em image registration using slice
  interpolation. In: IEEE ISBI (2014)

\bibitem{niklaus2017adaptive}
Niklaus, S., Mai, L., Liu, F.: Video frame interpolation via adaptive
  convolution. In: CVPR (2017)

\bibitem{niklaus2017separable}
Niklaus, S., Mai, L., Liu, F.: Video frame interpolation via adaptive separable
  convolution. In: ICCV (2017)

\bibitem{ranjan2017optical}
Ranjan, A., Black, M.J.: Optical flow estimation using a spatial pyramid
  network. In: CVPR (2017)

\bibitem{ronneberger2015u}
Ronneberger, O., Fischer, P., Brox, T.: U-net: Convolutional networks for
  biomedical image segmentation. In: International Conference on Medical image
  computing and computer-assisted intervention (2015)

\bibitem{roy2016monocular}
Roy, A., Todorovic, S.: Monocular depth estimation using neural regression
  forest. In: CVPR (2016)

\bibitem{wang2015towards}
Wang, P., Shen, X., Lin, Z., Cohen, S., Price, B., Yuille, A.L.: Towards
  unified depth and semantic prediction from a single image. In: CVPR (2015)

\bibitem{zhang2018self}
Zhang, H., Goodfellow, I., Metaxas, D., Odena, A.: Self-attention generative
  adversarial networks. arXiv:1805.08318  (2018)

\bibitem{zhang2018residual}
Zhang, Y., Tian, Y., Kong, Y., Zhong, B., Fu, Y.: Residual dense network for
  image restoration. arXiv preprint arXiv:1812.10477  (2018)

\end{thebibliography}
}

\end{document}